\documentclass[10pt,twocolumn,letterpaper]{article}

\usepackage{cvpr}              

\usepackage{booktabs}
\usepackage{multirow}
\usepackage[normalem]{ulem}
\useunder{\uline}{\ul}{}

\definecolor{cvprblue}{rgb}{0.21,0.49,0.74}
\usepackage[pagebackref,breaklinks,colorlinks,allcolors=cvprblue]{hyperref}
\usepackage{multirow}
\usepackage{array} 
\usepackage[table,xcdraw]{xcolor}
\usepackage{algorithm}
\usepackage{algpseudocode}

\usepackage{kotex}
\usepackage{comment}
 
\usepackage[accsupp]{axessibility}  

\def\paperID{} 
\def\confName{CVPR}
\def\confYear{2026}
 
\title{ G-MIXER: Geodesic Mixup-based Implicit Semantic Expansion and \\
Explicit Semantic Re-ranking for Zero-Shot Composed Image Retrieval
}

\author{
Jiyoung Lim \thanks{Equal contribution} \qquad Heejae Yang\footnotemark[1] \qquad Jee-Hyong Lee\thanks{Corresponding author.} \\
Sungkyunkwan University \\
{\tt\small \{maya0707, neulbo1126, john\}@skku.edu } \\
}

\begin{document}
\maketitle

\begin{abstract}
Composed Image Retrieval (CIR) aims to retrieve target images by integrating a reference image with a corresponding modification text. CIR requires jointly considering the explicit semantics specified in the query and the implicit semantics embedded within its bi-modal composition. Recent training-free Zero-Shot CIR (ZS-CIR) methods leverage Multimodal Large Language Models (MLLMs) to generate detailed target descriptions, converting the implicit information into explicit textual expressions. However, these methods rely heavily on the textual modality and fail to capture the fuzzy retrieval nature that requires considering diverse combinations of candidates. This leads to reduced diversity and accuracy in retrieval results. To address this limitation, we propose a novel training-free method, Geodesic Mixup-based Implicit semantic eXpansion and Explicit semantic Re-ranking for ZS-CIR (G-MIXER). G-MIXER constructs composed query features that reflect the implicit semantics of reference image-text pairs through geodesic mixup over a range of mixup ratios, and builds a diverse candidate set. The generated candidates are then re-ranked using explicit semantics derived from MLLMs, improving both retrieval diversity and accuracy. Our proposed G-MIXER achieves state-of-the-art performance across multiple ZS-CIR benchmarks, effectively handling both implicit and explicit semantics without additional training. Our code will be available at https://github.com/maya0395/gmixer.


\end{abstract}    
\section{Introduction}
\label{sec:intro}

\begin{figure*}[h!] 
\begin{center}
\includegraphics[width=1\linewidth]{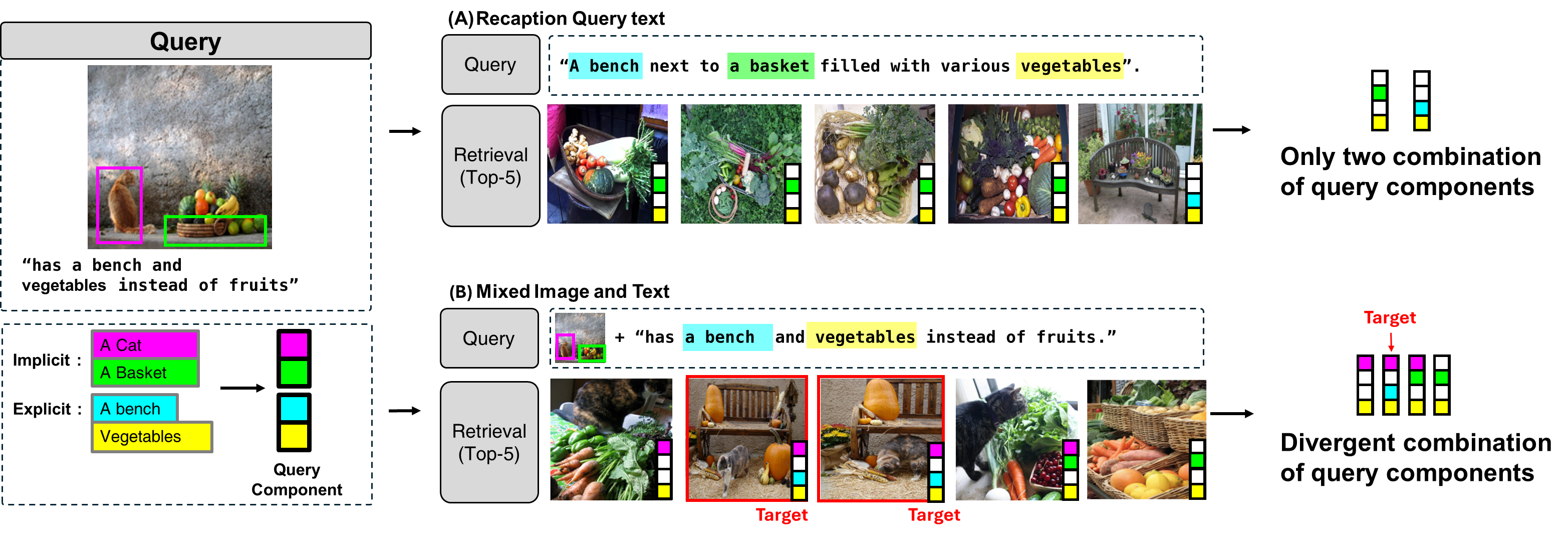}
\end{center}
\vspace{-1em}
\caption{\textbf{Illustration of our motivation.} The query contains \textbf{implicit} information \textit{cat, basket} present in the image but not mentioned in the text, while the modification text includes \textbf{explicit} attributes such as \textit{bench} and \textit{vegetables}. The MLLM based caption generation approach narrows the retrieval scope by converting implicit information from reference image into explicit descriptions. In contrast, the composed query  that incorporates information expands the retrieval scope and produces more diverse combinations of results.}
\label{fig:problem}
\vspace{-1em}
\end{figure*}

Composed Image Retrieval (CIR)~\cite{vo2019composing} addresses the challenge of integrating visual content and textual semantics to retrieve target images that reflect both the information in a reference image and the modifications specified by a user-provided text. By jointly leveraging the two modalities of image and text, CIR enables more fine-grained and intuitive retrieval.
This is achieved by combining attributes that are more naturally expressed in language with those that are more distinctive in the visual domain. However, supervised CIR methods require costly annotated triplets and suffer from poor generalization across domains.

To overcome these limitations, recent Zero-Shot CIR (ZS-CIR) methods~\cite{baldrati2023zero,saito2023pic2word,tang2024context} have been proposed, which do not require annotated triplet datasets. Existing ZS-CIR methods retrieve images using bi-modal queries, typically relying on the cross-modal alignment of Vision-Language Pretraining~(VLP) models such as CLIP~\cite{radford2021learning} and BLIP~\cite{li2022blip}.
Here, the core challenge lies in how to effectively handle the heterogeneous nature of bi-modal queries.
Specifically, bi-modal queries consist of a modification text and a reference image. Due to the nature of queries composed of different modalities, retrieval must account for both the implicit information contained solely in the image and the explicit information described in the text.

As illustrated in Figure~\ref{fig:problem}, the modification text \textbf{explicitly} specifies \textit{bench} and \textit{vegetables} to replace the fruits in the reference image. However, \textbf{implicit} elements such as the \textit{cat} and \textit{basket} are not clearly defined, making it ambiguous whether they should appear in the target image. 
As illustrated in (A), when an MLLM is used to generate a target-image description, the resulting caption primarily reflects explicit cues such as the bench and vegetables.
In contrast, (B) shows that using a composed query feature that fuses image and text information yields diverse retrieval results that include various combinations of elements. 
This observation indicates that implicit information makes it difficult to determine which elements should appear in the target image, highlighting the need for diversity in retrieval results. Therefore, it is crucial to perform fuzzy retrieval~\cite{ bordogna1993fuzzy, chen2002region, tahani1976fuzzy, yang2024ldre} to effectively compose ambiguous cross-modal queries.


This observation highlights that ZS-CIR requires jointly considering explicit semantics, conditions clearly stated in the modification text, and implicit semantics, visual elements present in the reference image but unmentioned in the text whose inclusion in the target image is inherently ambiguous. Effective retrieval depends on jointly reasoning over these two aspects, as users often describe their intent ambiguously, leaving certain visual details unstated. A variety of approaches have been explored to mitigate this ambiguity.
Some attempt to make implicit cues more explicit through reasoning or textual generation~\cite{tang2025reason, yang2024ldre}, while others directly combine image and text features to capture both modalities~\cite{jang2024spherical}. However, these approaches still tend to overemphasize one aspect, either precision from explicit reasoning or diversity from implicit blending, without fully integrating the two.

Consequently, ZS-CIR requires a fuzzy retrieval process that expands the retrieval scope to include diverse results, while filtering out noisy candidates through attributes explicitly expressed within the query. To achieve this, we propose Geodesic Mixup-based Implicit semantic eXpansion and Explicit semantic Re-ranking for Zero-Shot Composed Image Retrieval (G-MIXER), which incorporates as much implicit information as possible and refines retrieval results based on explicit cues. Specifically, G-MIXER consists of two complementary modules. Geodesic Mixup-based Implicit semantic eXpansion (G-MIX) expands the retrieval scope to cover diverse implicit semantics by varying the mixup ratio. We also propose Explicit semantic Re-ranking (ER) to remove noisy candidates by defining attributes explicitly expressed in the modification text. This approach enables the model to flexibly reason over both implicit and explicit information, achieving both diverse and precise retrieval results.

In summary, our contributions can be summarized as follows:
\begin{itemize}
    \item We propose a retrieval method that expands the search scope by composing queries along geodesic paths between image and text representations. This approach enables the model to capture diverse implicit semantics more effectively.
    \item We introduce a re-ranking strategy that refines retrieved candidates by leveraging explicit cues for filtering, effectively removing noisy or irrelevant results.
    \item Our method achieves state-of-the-art performance across multiple zero-shot CIR benchmarks, demonstrating its ability to jointly handle implicit and explicit semantics without additional training. 
\end{itemize}


\begin{figure*}[ t!] 
\begin{center}
\includegraphics[width=1\linewidth]{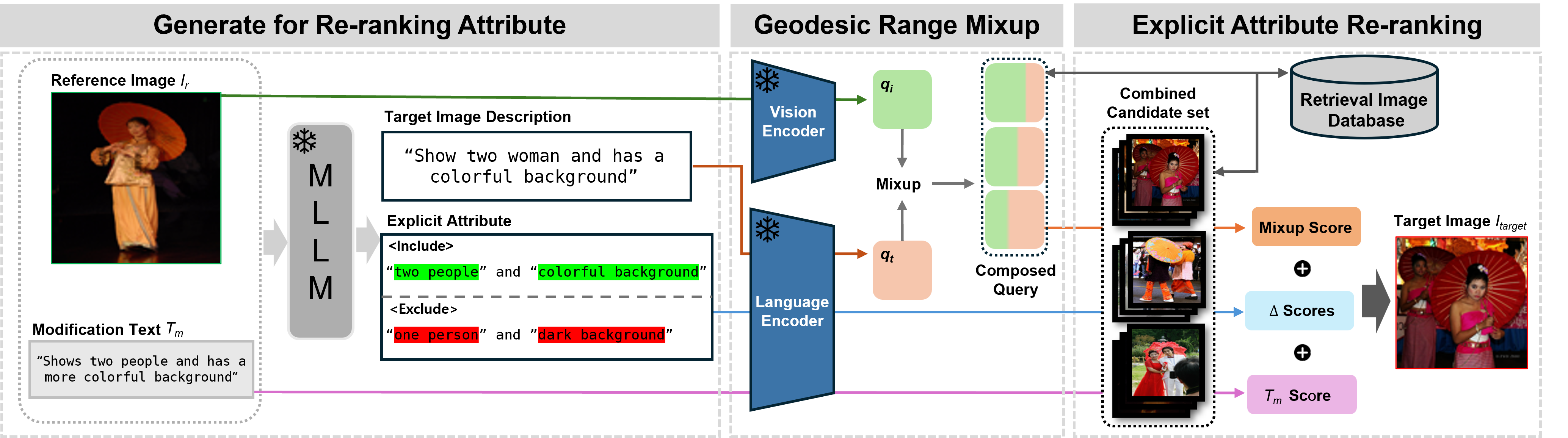}
\end{center}
\vspace{-1em} 
\caption{\textbf{Overview of G-MIXER.} (1) Generate the Target Image Description and Re-ranking Attributes using MLLM; (2) Perform extended Geodesic Mixup on the features obtained by encoding the Target Image Description and Reference Image with the pretrained CLIP encoder; (3) Filter noisy samples through Explicit Semantic Re-ranking. }
\label{fig:method}
\vspace{-1em}
\end{figure*}

\section{RELATED WORK}
\label{sec:related work}
\subsection{Composed Image Retrieval}
Composed Image Retrieval (CIR) is a task that retrieve a target image by combining a reference image with a modification text~\cite{vo2019composing}. It is a multimodal retrieval problem that requires the joint understanding of visual and textual conditions. Early supervised CIR methods train on image-text triplets to compose query representations~\cite{baldrati2022effective, liu2021image, psomas2025instance}, but such approaches are fundamentally constrained by their reliance on large scale annotated triplet data. Constructing such datasets requires substantial human effort and time. Moreover, their reliance on annotated data makes them inefficient in terms of scalability and generalization.

To mitigate these limitations, recent studies have focused on Zero-Shot Composed Image Retrieval (ZS-CIR)~\cite{karthik2023vision, baldrati2023zero, du2024image2sentence, gulanguage, jang2024spherical, lin2024fine, saito2023pic2word, suo2024knowledge, agnolucci2025isearle}. ZS-CIR aims to build models that perform CIR without annotated triplet data and can be categorized into training-based and training-free approaches. 
Training-based ZS-CIR methods~\cite{baldrati2023zero, lin2024fine, agnolucci2025isearle} learn to perform Textual Inversion from image-caption pairs. This approach employs the pretrained CLIP text encoder to project image features into the textual space. It then utilized the resulting pseudo word tokens to model the compositional relationship between images and text. 
Training-free ZS-CIR methods~\cite{tang2025reason, karthik2023vision, yang2024ldre} generate new target descriptions without any additional training by leveraging Multimodal Large Language Models (MLLMs). Representative approaches such as CIReVL~\cite{karthik2023vision} and OSrCIR~\cite{tang2025reason} first caption the reference image and then reconstruct a refined description that reflects the modification text through a MLLM.

However, these methods primarily rely on textual reasoning,
which results in a narrow retrieval scope and a text-dominant interpretation of compositional queries.
Such reliance makes it difficult to \textbf{effectively capture the compositional interaction between explicit and implicit information}. To address these limitations, we propose G-MIXER Geodesic Mixup-based Implicit semantic eXpansion and Explicit semantic Re-ranking. G-MIXER performs fuzzy retrieval by incorporating implicit semantics through an expanded Geodesic Mixup, and enhances explicit representation through an Explicit semantic Re-ranking process. 
\subsection{Vision and Language Pre-training Model}
Vision-Language Pretraining Model (VLP) is a pretrained model that learns semantic alignment between images and texts using large-scale image-text pairs~\cite{jia2021scaling, radford2021learning, yao2021filip}. A representative example is Contrastive Language-Image Pre-training CLIP~\cite{radford2021learning}, which jointly trains an image encoder and a text encoder to align the representation spaces of the two modalities, thereby effectively bridging visual concepts and linguistic expressions. The pretrained VLPs~\cite{radford2021learning, li2022blip} can be applied to various multimodal tasks in a zero-shot manner~\cite{alayrac2022flamingo, hummel2024egocvr, jiang2023clip}, demonstrating strong generalization capability in applications such as image–text retrieval and vision–language understanding.

Based on these VLP encoders, research on Multimodal Large Language Models (MLLMs) has been actively conducted in recent years~\cite{li2023blip, bai2023qwen, liu2024improved}. While conventional VLPs focus on aligning visual and textual representations, Multimodal Large Language Models (MLLMs) extend this capability by integrating the reasoning ability of language models. This integration enables deeper multimodal understanding and generation. For example, models such as LLaVA~\cite{liu2023visual}, MiniGPT-4~\cite{zhu2023minigpt}, and GPT-4V(ision)~\cite{achiam2023gpt} connect a pretrained vision encoder to a large language model (LLM), enabling the interpretation of visual information in linguistic form and visual reasoning conditioned on text.
These models go beyond simple alignment learning, expanding their applications to text-conditioned generation, visual question answering (VQA), and compositional image understanding~\cite{peng2023kosmos, dai2023instructblip, liu2024improved, wu2024visionllm}.

In our work, we demonstrate that ZS-CIR can be performed efficiently and scalably without additional training by integratively leveraging pretrained VLP encoders and MLLMs.


\section{Method}
The goal of Zero-Shot Composed Image Retrieval (ZS-CIR) is to retrieve a target image $I_{t}$ from an image database $D$ that is visually similar to the reference image $I_r$ while reflecting the modifications specified in the modification text $T_m$. Training-free approaches generate a target description $T_t$ to retrieve the target image $I_{t}$ by measuring the similarity between target description $T_{t}$ and candidate images in $D$. The similarity score $S$ is computed using cosine similarity $\cos(\Psi_I(I_c), \Psi_T(T_{t}))$, where $\Psi_I$ and $\Psi_T$ denote the image and text encoders of pretrained CLIP, respectively. 

The query in ZS-CIR inherently involves both \textbf{explicit} information, which specifies what should change and \textbf{implicit} information which is not directly expressed. Because of these implicit semantics, it becomes unclear which elements should be explicitly transformed for retrieval, making the query inherently ambiguous. If all implicit information is converted into explicit form, the retrieval space becomes overly constrained. 
Conversely, partial conversion of implicit information may cause the loss of important cues and degrade retrieval performance. To address this challenge, we propose Geodesic Mixup-based Implicit semantic eXpansion retrieval and Explicit semantic Re-ranking (G-MIXER) method for ZS-CIR. G-MIXER constructs an expanded retrieval space that preserves implicit semantics and refines noisy candidates using explicit semantic information. An overview of the proposed method is shown in Figure~\ref{fig:method}.

Our method consist of two stages.\\
(1) \textbf{Geodesic Mixup-based Implicit semantic eXpansion Retrieval}: Composed query features are generated by applying geodesic mixup over a range of mixup ratios to capture implicit semantics and to build a diverse candidate set.
(Sec.~\ref{sec:mixup}).
\\
(2) \textbf{Explicit semantic Re-ranking}: Noisy candidates are filtered out by re-ranking the results using explicit semantics(\texttt{Include}/\texttt{Exclude}) generated by MLLMs.
(Sec.~\ref{sec:rerank})
. 

\begin{table*}[t!]    
\centering
\scriptsize
\begin{tabular}{@{}cc | cccc | cccccc@{}}
\toprule
\multicolumn{2}{c}{\textbf{Datasets} } & \multicolumn{4}{|c}{\textbf{CIRCO}}  
& \multicolumn{6}{|c}{\textbf{CIRR}} \\ \midrule
\multicolumn{2}{c|}{Metric}   
& \multicolumn{4}{c|}{mAP@k}  
& \multicolumn{3}{c|}{Recall@k} 
& \multicolumn{3}{c}{ $\text{Recall}_{\text{Subset}}$@k  }   \\
\multicolumn{1}{c|}{Arch}                
& \multicolumn{1}{c|}{Method}                                
& k=5  & k=10 & k=25   
& \multicolumn{1}{c|}{k=50} 
& k=1  & k=5 
& \multicolumn{1}{c|}{k=10} & k=1   & k=2 & k=3 \\ 
\midrule
\multicolumn{1}{c|}{}  
& \multicolumn{1}{c|}{SEARLE} 
& 9.35 & 9.94  & 11.13 & \multicolumn{1}{c|}{11.84}   
& 24.00 & 53.42 & \multicolumn{1}{c|}{66.82} & 54.89 & 76.60  & 88.19   \\

\multicolumn{1}{c|}{} 
& \multicolumn{1}{c|}{\cellcolor[HTML]{EFEFEF}CIReVL}        
& \cellcolor[HTML]{EFEFEF}14.94       
& \cellcolor[HTML]{EFEFEF}15.42       
& \cellcolor[HTML]{EFEFEF}17.00       
& \multicolumn{1}{c|}{\cellcolor[HTML]{EFEFEF}17.82}       
& \cellcolor[HTML]{EFEFEF}23.94       
& \cellcolor[HTML]{EFEFEF}52.51       
& \multicolumn{1}{c|}{\cellcolor[HTML]{EFEFEF}66.00}       
& \cellcolor[HTML]{EFEFEF}60.17       
& \cellcolor[HTML]{EFEFEF}80.05       
& \cellcolor[HTML]{EFEFEF}90.19 \\

\multicolumn{1}{c|}{}  
& \multicolumn{1}{c|}{\cellcolor[HTML]{EFEFEF}LDRE}         
& \cellcolor[HTML]{EFEFEF}17.96       
& \cellcolor[HTML]{EFEFEF}18.32       
& \cellcolor[HTML]{EFEFEF}20.21       
& \multicolumn{1}{c|}{\cellcolor[HTML]{EFEFEF}21.11}       
& \cellcolor[HTML]{EFEFEF}{\ul 25.69} 
& \cellcolor[HTML]{EFEFEF}{\ul 55.13} 
& \multicolumn{1}{c|}{\cellcolor[HTML]{EFEFEF}{\ul 69.04}} 
& \cellcolor[HTML]{EFEFEF}60.53       
& \cellcolor[HTML]{EFEFEF}80.65       
& \cellcolor[HTML]{EFEFEF}90.70 \\

\multicolumn{1}{c|}{}  & \multicolumn{1}{c|}{\cellcolor[HTML]{EFEFEF}OSrCIR}        
& \cellcolor[HTML]{EFEFEF}{\ul 18.04} 
& \cellcolor[HTML]{EFEFEF}{\ul 19.17} 
& \cellcolor[HTML]{EFEFEF}{\ul 20.94} 
& \multicolumn{1}{c|}{\cellcolor[HTML]{EFEFEF}{\ul 21.85}} 
& \cellcolor[HTML]{EFEFEF}25.42       
& \cellcolor[HTML]{EFEFEF}54.54       
& \multicolumn{1}{c|}{\cellcolor[HTML]{EFEFEF}68.19}       
& \cellcolor[HTML]{EFEFEF}{\ul 62.31} 
& \cellcolor[HTML]{EFEFEF}{\ul 80.86} 
& \cellcolor[HTML]{EFEFEF}{\ul 91.13} \\

\multicolumn{1}{c|}{\multirow{-5}{*}{ViT-B/32}} 
& \multicolumn{1}{c|}{\cellcolor[HTML]{EFEFEF}\textbf{ G-MIXER~(Ours) }} 
& \cellcolor[HTML]{EFEFEF} \textbf{21.33}
& \cellcolor[HTML]{EFEFEF} \textbf{22.00}
& \cellcolor[HTML]{EFEFEF}   \textbf{24.00}
& \multicolumn{1}{c|}{\cellcolor[HTML]{EFEFEF}{ \textbf{24.98} }}            
& \cellcolor[HTML]{EFEFEF}    \textbf{35.18}
& \cellcolor[HTML]{EFEFEF}   \textbf{66.19}
& \multicolumn{1}{c|}{\cellcolor[HTML]{EFEFEF}{ \textbf{76.82} }}            
& \cellcolor[HTML]{EFEFEF}   \textbf{69.98}        
& \cellcolor[HTML]{EFEFEF}   \textbf{85.78}       
& \cellcolor[HTML]{EFEFEF}   \textbf{ 92.92 }        \\ \midrule

\multicolumn{1}{c|}{}                           
& \multicolumn{1}{c|}{SEARLE}                                
& 11.68   & 12.73  & 14.33  & \multicolumn{1}{c|}{15.12}  
& 24.24 & 52.48 & \multicolumn{1}{c|}{66.29} 
& 53.76  & 75.01  & 88.19 \\

\multicolumn{1}{c|}{} & \multicolumn{1}{c|}{LinCIR} 
& 12.59 & 13.58 & 15.00   & \multicolumn{1}{c|}{15.85}                               
& 25.04  & 53.25 & \multicolumn{1}{c|}{66.68} 
& 57.11  & 77.37 & 88.89  \\

\multicolumn{1}{c|}{}  & \multicolumn{1}{c|}{PrediCIR}                              
& 15.70  & 17.10  & 18.60  & \multicolumn{1}{c|}{19.30}                               
& 27.20   & 57.00  & \multicolumn{1}{c|}{70.20} 
& -  & -  & -   \\

\multicolumn{1}{c|}{} & \multicolumn{1}{c|}{\cellcolor[HTML]{EFEFEF}CIReVL}        
& \cellcolor[HTML]{EFEFEF}18.57       
& \cellcolor[HTML]{EFEFEF}19.01       
& \cellcolor[HTML]{EFEFEF}20.89       
& \multicolumn{1}{c|}{\cellcolor[HTML]{EFEFEF}21.80}       
& \cellcolor[HTML]{EFEFEF}24.55       
& \cellcolor[HTML]{EFEFEF}52.31       
& \multicolumn{1}{c|}{\cellcolor[HTML]{EFEFEF}64.92}       
& \cellcolor[HTML]{EFEFEF}59.54       
& \cellcolor[HTML]{EFEFEF}79.88       
& \cellcolor[HTML]{EFEFEF}89.69       \\

\multicolumn{1}{c|}{} & \multicolumn{1}{c|}{\cellcolor[HTML]{EFEFEF}LDRE}          
& \cellcolor[HTML]{EFEFEF}23.35       
& \cellcolor[HTML]{EFEFEF}24.03      
& \cellcolor[HTML]{EFEFEF}26.44       
& \multicolumn{1}{c|}{\cellcolor[HTML]{EFEFEF}27.50}       
& \cellcolor[HTML]{EFEFEF}26.53       
& \cellcolor[HTML]{EFEFEF}55.57       
& \multicolumn{1}{c|}{\cellcolor[HTML]{EFEFEF}67.54}       
& \cellcolor[HTML]{EFEFEF}60.43       
& \cellcolor[HTML]{EFEFEF}80.31       
& \cellcolor[HTML]{EFEFEF}89.90       \\

\multicolumn{1}{c|}{} & \multicolumn{1}{c|}{\cellcolor[HTML]{EFEFEF}OSrCIR}        
& \cellcolor[HTML]{EFEFEF}{\ul 23.87} 
& \cellcolor[HTML]{EFEFEF}{\ul 25.33} 
& \cellcolor[HTML]{EFEFEF}{\ul 27.84} 
& \multicolumn{1}{c|}{\cellcolor[HTML]{EFEFEF}{\ul 28.97}} 
& \cellcolor[HTML]{EFEFEF}{\ul 29.45} 
& \cellcolor[HTML]{EFEFEF}{\ul 57.68} 
& \multicolumn{1}{c|}{\cellcolor[HTML]{EFEFEF}{\ul 69.86}} 
& \cellcolor[HTML]{EFEFEF}{\ul 62.12} 
& \cellcolor[HTML]{EFEFEF}81.92       
& \cellcolor[HTML]{EFEFEF}{\ul 91.10} \\
 
\multicolumn{1}{c|}{\multirow{-7}{*}{ViT-L/14}} 
& \multicolumn{1}{c|}{\cellcolor[HTML]{EFEFEF}\textbf{G-MIXER~(Ours) }} 
& \cellcolor[HTML]{EFEFEF}   \textbf{28.29}
& \cellcolor[HTML]{EFEFEF}   \textbf{29.04}
& \cellcolor[HTML]{EFEFEF}   \textbf{31.44}
& \multicolumn{1}{c|}{\cellcolor[HTML]{EFEFEF} { \textbf{32.39}} }     
& \cellcolor[HTML]{EFEFEF}      \textbf{37.42}
& \cellcolor[HTML]{EFEFEF}   \textbf{67.69}
& \multicolumn{1}{c|}{\cellcolor[HTML]{EFEFEF}{ \textbf{78.58}} }     
& \cellcolor[HTML]{EFEFEF}    \textbf{71.88}        
& \cellcolor[HTML]{EFEFEF}    \textbf{87.04}        
& \cellcolor[HTML]{EFEFEF}    \textbf{92.82}        \\ \midrule

\multicolumn{1}{c|}{} & \multicolumn{1}{c|}{LinCIR} 
& 19.71 & 21.01 & 23.13 & \multicolumn{1}{c|}{24.18} 
& 35.25 & 64.72 & \multicolumn{1}{c|}{76.05} 
& 63.35 & 82.22 & 91.98 \\

\multicolumn{1}{c|}{} & \multicolumn{1}{c|}{PrediCIR}
& 23.70 & 24.60 & 25.40 & \multicolumn{1}{c|}{26.00} 
& 37.00 & 66.10 & \multicolumn{1}{c|}{{\ul 77.90}} 
& {\ul -} & - & - \\

\multicolumn{1}{c|}{} 
& \multicolumn{1}{c|}{\cellcolor[HTML]{EFEFEF}CIReVL}        
& \cellcolor[HTML]{EFEFEF}26.77       
& \cellcolor[HTML]{EFEFEF}27.59       
& \cellcolor[HTML]{EFEFEF}29.96       
& \multicolumn{1}{c|}{\cellcolor[HTML]{EFEFEF}31.03}       
& \cellcolor[HTML]{EFEFEF}34.65       
& \cellcolor[HTML]{EFEFEF}64.29       
& \multicolumn{1}{c|}{\cellcolor[HTML]{EFEFEF}75.06}       
& \cellcolor[HTML]{EFEFEF}67.95       
& \cellcolor[HTML]{EFEFEF}84.87       
& \cellcolor[HTML]{EFEFEF}93.21       \\

\multicolumn{1}{c|}{} & \multicolumn{1}{c|}{\cellcolor[HTML]{EFEFEF}LDRE}
& \cellcolor[HTML]{EFEFEF}{\ul 31.12} 
& \cellcolor[HTML]{EFEFEF}{\ul 32.24} 
& \cellcolor[HTML]{EFEFEF}34.95       
& \multicolumn{1}{c|}{\cellcolor[HTML]{EFEFEF}36.03}       
& \cellcolor[HTML]{EFEFEF}36.15       
& \cellcolor[HTML]{EFEFEF}66.39       
& \multicolumn{1}{c|}{\cellcolor[HTML]{EFEFEF}77.25}       
& \cellcolor[HTML]{EFEFEF}68.82       
& \cellcolor[HTML]{EFEFEF}85.66       
& \cellcolor[HTML]{EFEFEF} {\ul 93.76 }  \\

\multicolumn{1}{c|}{} & \multicolumn{1}{c|}{\cellcolor[HTML]{EFEFEF}OSrCIR}
& \cellcolor[HTML]{EFEFEF}30.47       
& \cellcolor[HTML]{EFEFEF}31.14       
& \cellcolor[HTML]{EFEFEF}{\ul 35.03} 
& \multicolumn{1}{c|}{\cellcolor[HTML]{EFEFEF}{\ul 36.59}} 
& \cellcolor[HTML]{EFEFEF}{\ul 37.26} 
& \cellcolor[HTML]{EFEFEF}{\ul 67.25} 
& \multicolumn{1}{c|}{\cellcolor[HTML]{EFEFEF}77.33}       
& \cellcolor[HTML]{EFEFEF}{\ul 69.22} 
& \cellcolor[HTML]{EFEFEF}{\ul 85.28} 
& \cellcolor[HTML]{EFEFEF}  93.55         \\

\multicolumn{1}{c|}{\multirow{-6}{*}{ViT-G/14}} 
& \multicolumn{1}{c|}{\cellcolor[HTML]{EFEFEF}\textbf{G-MIXER~(Ours)}}
& \cellcolor[HTML]{EFEFEF} \textbf{31.79}
& \cellcolor[HTML]{EFEFEF} \textbf{32.54}
& \cellcolor[HTML]{EFEFEF} \textbf{35.49}
& \cellcolor[HTML]{EFEFEF} \textbf{36.87}

& \cellcolor[HTML]{EFEFEF} \textbf{39.18}
& \cellcolor[HTML]{EFEFEF} \textbf{69.83}
& \multicolumn{1}{c|}{\cellcolor[HTML]{EFEFEF} \textbf{79.35} }   

& \cellcolor[HTML]{EFEFEF} \textbf{72.36}
& \cellcolor[HTML]{EFEFEF} \textbf{87.25}
& \cellcolor[HTML]{EFEFEF} \textbf{93.85} \\ \bottomrule
\end{tabular}
\label{tab:circo}
\caption{\textbf{Comparison on CIRCO and CIRR Test Data.} 
G-MIXER outperforms existing ZS-CIR methods on both datasets CIRCO and CIRR. Gray rows indicate training-free ZS-CIR methods. \textbf{Bold} and \underline{underlined} values denote the best and second-best results, respectively.}
\end{table*}


\subsection{Geodesic Mixup-based Implicit semantic 
eXpansion Retrieval (G-MIX Retrieval)}
\label{sec:mixup}

To expand the retrieval scope by capturing \textbf{implicit information} embedded in the query, we propose Implicit Semantic Expansion Retrieval based on Geodesic Mixup (G-MIX). Since the CLIP embedding space forms a hypersphere based on cosine similarity, simple linear mixup can distort the underlying geometry. Therefore, we apply geodesic mixup which follows the shortest path on the hypersphere. 
We perform geodesic mixup between the image feature $f_i = \Psi_I(I_r)$ and the text feature $f_t = \Psi_T(T_m)$ from the given reference image–text pair $(I_r, T_m)$ to obtain a composed query $m_{\lambda}$ that integrates both implicit and explicit semantics. 
The resulting mixed feature is subsequently used as the query representation to retrieve candidate images. Formally, the composed query feature is defined as:
 
\begin{equation}
    m_{\lambda}(f_t, f_i) 
= f_t \frac{\sin(\lambda \theta)}{\sin(\theta)} 
+ f_i \frac{\sin((1 - \lambda)\theta)}{\sin(\theta)}, 
\label{eq:mixup}
\end{equation}
where $\lambda$ denotes the mixing ratio, and $\theta = \cos^{-1}(f_t \cdot f_i) $ represents the angle between the two features. 
For example, $m_{0.8}(f_t, f_i)$ denotes a composed query feature in which the text feature is weighted by 0.8 and the image feature is weighted by 0.2. 
For smaller values of $\lambda$, structural and background information from the image are more strongly reflected, whereas for larger $\lambda$, the attribute changes specified by the text become more prominent.

Changing the mixing ratio $\lambda$ creates a continuous semantic trajectory that smoothly interpolates between the image and text modalities.
We generate multiple composed queries along this trajectory and perform retrieval for each. The retrieved results with high similarity scores are aggregated to form a candidate set. 
Instead of searching the entire embedding space indiscriminately,
this approach samples along the principal semantic axis to cover dense semantic transitions with only $N$ mixing ratios. 
This enables a stable and effective expansion of the retrieval scope. Retrieval is performed for each composed query $m_\lambda(f_t, f_i)$, selecting the top-$K$ images according to cosine similarity $S_{\lambda}$ :
\begin{equation}
\text{Retrieval}\!\left(m_{\lambda}(f_t, f_i), D \right)
= 
\operatorname{TopK}\big( s_{\lambda}, k \big). 
\label{eq:topk}
 \end{equation}
 
For example, when $\lambda$ ranges from 0.7 to 1.0 with a step of 0.1 ($N=4$ mixing ratios), we retrieve the top-100 images for each $m_{\lambda}$ and aggregate them into a set of 400 first-stage candidates (Eq.~\ref{eq:topk}). Because higher $\lambda$ values naturally yield higher similarity scores, we apply min–max normalization to scale all scores between 0 and 1.
For candidates retrieved at multiple ratios, the maximum score is used to construct the final set.

\begin{equation}
\mathcal{R}_{\text{union}} = \bigcup_{\lambda \in [0.7, 1] }    
\text{Retrieval}\bigg(m_{\lambda} \big(  f_t, f_i \big) , D \bigg).
\label{eq:union}
\end{equation}
We used $T_{t}$, a target description generated by MLLM based on $T_{m}$ and the reference image $I_r$. Here, $T_{t}$ is not a reasoning-based recaption, but rather a caption generated by referring to the reference image $I_r$ to supplement omitted subjects or comparative expressions in $T_{m}$.

As a result, our G-MIX Retrieval
alleviates semantic ambiguity between image and text modalities and effectively expands the compositional semantic scope by leveraging Geodesic Mixup.

\subsection{Explicit semantic Re-ranking (ER)}
\label{sec:rerank}
To remove noisy candidates included in the expanded retrieval set, we propose Explicit semantic Re-ranking (ER), which leverages explicit cues for refinement. Since previous re-ranking methods rely on multiple captions generated by MLLMs, the captions inevitably contain implicit information. However, since these implicit cues do not clearly indicate whether they should appear in the target image, even higher similarity scores cannot be regarded as a reliable basis for ranking. To address this issue, ER uses \textbf{explicit information} extracted by MLLMs as the basis for re-ranking. For each image $I_{set}$ included in the first-stage retrieval set $\mathcal{R}_\text{union}$, we compute the similarity $S_{\lambda} = \cos\big(m_{\lambda}, \Psi_I(I_{set})\big)$ with each $m_{\lambda}$. Based on the variation in similarity with explicit attributes, we adjust the ranking to prioritize candidates that include clearer information and filter out noisy samples dominated by implicit cues. 

To convert the explicit information in the query into a caption, we design a prompt $p_s$. 
We generate the \texttt{Include} and \texttt{Exclude} texts based on the query pair $(I_r, T_m)$ to ensure that the explicitly defined conditions are accurately captured in the retrieval process. 

Formally, the generation process using the MLLM $\Psi_M$ is defined as:
\begin{equation}
    T_{t}, T_{in}, T_{ex} = \Psi_M (p_s \circ I_r \circ T_m),
\end{equation}
where $\circ$ denotes the concatenation of inputs.

To achieve explicit semantic re-ranking, we measure the similarities $S_{in}$ and $S_{ex}$ between each image $I_{set}$ and $T_{in}$ and $T_{ex}$, respectively. 
Intuitively, a candidate image should have a high similarity to the explicit include caption $T_{in}$ while maintaining a low similarity to the explicit exclude caption $T_{ex}$.
We capture this behavior by comparing $S_{in}$ and $S_{ex}$ against $S_{\lambda}$.
When $S_{\lambda} - S_{in} $ is large, the candidate is closer to the composed query than to the include caption, suggesting that the desired attribute is not clearly represented, which acts as a penalty.
Conversely, when $S_{\lambda} - S_{ex} $ is large, the candidate is much less similar to the exclude caption than to the composed query, indicating that the undesired attribute is effectively suppressed, which serves as a reward.
These effects are summarized in the similarity difference $\Delta$, defined as:

\begin{equation}
\Delta =   ReLU( S_{\lambda} - S_{ex}  ) - ReLU( S_{\lambda} - S_{in} ) 
\label{eq:delta}
\end{equation}
where a larger $\Delta$ favors candidates that better satisfy the explicit include condition while avoiding the explicit exclude condition.

The final score is computed by combining the similarity $S_m$ obtained from the modification text with the similarity differences $\Delta$ derived from $S_{in}$ and $S_{ex}$ based on the MLLM-generated captions. 
\begin{equation}
\text{Final score} =  S_m + S_{\lambda} +  \Delta
\label{eq:final}
\end{equation}
This score assesses whether the retrieved candidates, which capture diverse implicit semantics through G-MIX, also include the intended explicit information. By re-ranking based on explicit semantic cues, our method effectively filters out noisy candidates while preserving diversity in the retrieval results. Details regarding the prompt formulation and other implementation specifics are described in Appendix.


\begin{table*}[t!]
\centering   
\scriptsize
\begin{tabular}{@{}cc|cccccc|cc@{}}
\toprule

\multicolumn{2}{c|}{\textbf{Fashion-IQ}} 
& \multicolumn{2}{c}{Shirt} & \multicolumn{2}{c}{Dress} & \multicolumn{2}{c|}{Toptee} 
& \multicolumn{2}{c}{\textbf{Average}} \\ \midrule

\multicolumn{1}{c|}{Backbone} & Method 
& R@10 & R@50 & R@10 & R@50 & R@10 & R@50 & R@10 & R@50 \\ \midrule
\multicolumn{1}{c|}{} & SEARLE & 24.44 & 41.61 & 18.54 & 39.51 & 25.70 & 46.46 & 22.89 & 42.53 \\

\multicolumn{1}{c|}{} & \cellcolor[HTML]{EFEFEF}CIReVL 
& \cellcolor[HTML]{EFEFEF}28.36 
& \cellcolor[HTML]{EFEFEF}47.84 
& \cellcolor[HTML]{EFEFEF}25.29 
& \cellcolor[HTML]{EFEFEF}46.36 
& \cellcolor[HTML]{EFEFEF}31.21 
& \cellcolor[HTML]{EFEFEF}53.85 
& \cellcolor[HTML]{EFEFEF}28.29 
& \cellcolor[HTML]{EFEFEF}49.35 \\

\multicolumn{1}{c|}{} & \cellcolor[HTML]{EFEFEF}LDRE          
& \cellcolor[HTML]{EFEFEF}27.38
& \cellcolor[HTML]{EFEFEF}46.27
& \cellcolor[HTML]{EFEFEF}19.97
& \cellcolor[HTML]{EFEFEF}41.84
& \cellcolor[HTML]{EFEFEF}27.07
& \cellcolor[HTML]{EFEFEF}48.78
& \cellcolor[HTML]{EFEFEF}24.81
& \cellcolor[HTML]{EFEFEF}45.63 \\
\multicolumn{1}{c|}{} & \cellcolor[HTML]{EFEFEF}OSrCIR 
& \cellcolor[HTML]{EFEFEF}{\ul 31.16} 
& \cellcolor[HTML]{EFEFEF}{\ul 51.13}
& \cellcolor[HTML]{EFEFEF}{\ul 29.35}
& \cellcolor[HTML]{EFEFEF}{\ul 50.37}
& \cellcolor[HTML]{EFEFEF}{\ul 36.51}
& \cellcolor[HTML]{EFEFEF}{\ul 58.71}
& \cellcolor[HTML]{EFEFEF}{\ul 32.34}
& \cellcolor[HTML]{EFEFEF}{\ul 53.40} \\

\multicolumn{1}{c|}{\multirow{-5}{*}{ViT-B/32}} 
& \cellcolor[HTML]{EFEFEF}\textbf{G-MIXER~(Ours)} 
& \cellcolor[HTML]{EFEFEF}\textbf{37.24} 
& \cellcolor[HTML]{EFEFEF}\textbf{55.99} 
& \cellcolor[HTML]{EFEFEF}\textbf{36.39} 
& \cellcolor[HTML]{EFEFEF}\textbf{58.21} 
& \cellcolor[HTML]{EFEFEF}\textbf{45.23} 
& \cellcolor[HTML]{EFEFEF}\textbf{64.10} 
& \cellcolor[HTML]{EFEFEF}\textbf{39.62} 
& \cellcolor[HTML]{EFEFEF}\textbf{59.43} \\ \midrule

\multicolumn{1}{c|}{} & SEARLE 
& 26.89 & 45.48 & 20.48 & 43.13 & 29.32 & 49.97 & 25.56 & 46.23 \\
\multicolumn{1}{c|}{} & PrediCIR & 31.80 & 52.00 & 25.40 & 49.50 & 33.10 & 55.40 & 30.10 & 52.30 \\

\multicolumn{1}{c|}{} & \cellcolor[HTML]{EFEFEF}CIReVL 
& \cellcolor[HTML]{EFEFEF}{29.49} 
& \cellcolor[HTML]{EFEFEF}{47.40} 
& \cellcolor[HTML]{EFEFEF}{24.79} 
& \cellcolor[HTML]{EFEFEF}{44.76} 
& \cellcolor[HTML]{EFEFEF}{31.36} 
& \cellcolor[HTML]{EFEFEF}{53.63} 
& \cellcolor[HTML]{EFEFEF}{28.55} 
& \cellcolor[HTML]{EFEFEF}{48.57} \\

\multicolumn{1}{c|}{} & \cellcolor[HTML]{EFEFEF}LDRE & \cellcolor[HTML]{EFEFEF}31.04 & \cellcolor[HTML]{EFEFEF}51.22 & \cellcolor[HTML]{EFEFEF}22.93 & \cellcolor[HTML]{EFEFEF}46.76 & \cellcolor[HTML]{EFEFEF}31.57 & \cellcolor[HTML]{EFEFEF}53.64 & \cellcolor[HTML]{EFEFEF}28.51 & \cellcolor[HTML]{EFEFEF}50.54 \\

\multicolumn{1}{c|}{} & \cellcolor[HTML]{EFEFEF}OSrCIR 
& \cellcolor[HTML]{EFEFEF}{\ul 33.17}
& \cellcolor[HTML]{EFEFEF}{\ul 52.03}
& \cellcolor[HTML]{EFEFEF}{\ul 29.70}
& \cellcolor[HTML]{EFEFEF}{\ul 51.81}
& \cellcolor[HTML]{EFEFEF}{\ul 36.92}
& \cellcolor[HTML]{EFEFEF}{\ul 59.27}
& \cellcolor[HTML]{EFEFEF}{\ul 33.26}
& \cellcolor[HTML]{EFEFEF}{\ul 54.37} \\

\multicolumn{1}{c|}{\multirow{-6}{*}{ViT-L/14}} & \cellcolor[HTML]{EFEFEF}
\textbf{G-MIXER~(Ours)} 
& \cellcolor[HTML]{EFEFEF}\textbf{40.87} 
& \cellcolor[HTML]{EFEFEF}\textbf{60.35} 
& \cellcolor[HTML]{EFEFEF}\textbf{37.98} 
& \cellcolor[HTML]{EFEFEF}\textbf{60.93} 
& \cellcolor[HTML]{EFEFEF}\textbf{46.91} 
& \cellcolor[HTML]{EFEFEF}\textbf{66.14} 
& \cellcolor[HTML]{EFEFEF}\textbf{41.92} 
& \cellcolor[HTML]{EFEFEF}\textbf{62.47} \\ \midrule

\multicolumn{1}{c|}{} & PrediCIR & \textbf{48.20} & \textbf{67.40} & \textbf{39.70} & \textbf{62.40} & \textbf{53.70} & \textbf{73.60} & \textbf{47.20} & \textbf{67.80} \\

\multicolumn{1}{c|}{} & \cellcolor[HTML]{EFEFEF}CIReVL 
& \cellcolor[HTML]{EFEFEF}33.71
& \cellcolor[HTML]{EFEFEF}51.42
& \cellcolor[HTML]{EFEFEF}27.07
& \cellcolor[HTML]{EFEFEF}49.53
& \cellcolor[HTML]{EFEFEF}35.80
& \cellcolor[HTML]{EFEFEF}56.14
& \cellcolor[HTML]{EFEFEF}32.19
& \cellcolor[HTML]{EFEFEF}52.36 \\

\multicolumn{1}{c|}{} & \cellcolor[HTML]{EFEFEF}LDRE 
& \cellcolor[HTML]{EFEFEF}35.94
& \cellcolor[HTML]{EFEFEF}58.58
& \cellcolor[HTML]{EFEFEF}26.11
& \cellcolor[HTML]{EFEFEF}51.12
& \cellcolor[HTML]{EFEFEF}35.42
& \cellcolor[HTML]{EFEFEF}56.67
& \cellcolor[HTML]{EFEFEF}32.49
& \cellcolor[HTML]{EFEFEF}55.46 \\

\multicolumn{1}{c|}{} & \cellcolor[HTML]{EFEFEF}OSrCIR
& \cellcolor[HTML]{EFEFEF}38.65
& \cellcolor[HTML]{EFEFEF}54.71
& \cellcolor[HTML]{EFEFEF}33.02
& \cellcolor[HTML]{EFEFEF}54.78
& \cellcolor[HTML]{EFEFEF}41.04
& \cellcolor[HTML]{EFEFEF}61.83
& \cellcolor[HTML]{EFEFEF}37.57
& \cellcolor[HTML]{EFEFEF}57.11 \\

\multicolumn{1}{c|}{\multirow{-5}{*}{ViT-G/14}} 
& \cellcolor[HTML]{EFEFEF}\textbf{G-MIXER~(Ours)} 
& \cellcolor[HTML]{EFEFEF}{\ul 39.65}    
& \cellcolor[HTML]{EFEFEF}{\ul 59.61}    
& \cellcolor[HTML]{EFEFEF}{\ul 34.71}
& \cellcolor[HTML]{EFEFEF}{\ul 58.85}
& \cellcolor[HTML]{EFEFEF}{\ul 44.77}
& \cellcolor[HTML]{EFEFEF}{\ul 67.47}
& \cellcolor[HTML]{EFEFEF}{\ul 39.71}
& \cellcolor[HTML]{EFEFEF}{\ul 61.98}    \\ 
\bottomrule
\end{tabular}
\label{tab:fashion}
\caption{\textbf{Comparison on FashionIQ Validation set.} G-MIXER is able to significantly outperform existing training-free ZS-CIR methods across all sub-benchmarks. Gray rows indicate training-free ZS-CIR methods. \textbf{Bold} and \underline{underlined} values denote the best and second-best results, respectively.}
\vspace{-1em}
\end{table*}

\begin{table}[h!]
\centering
\scriptsize
\begin{tabular}{@{}ccccc@{}}
\toprule
\multicolumn{5}{c}{\textbf{GeneCIS}} \\ \midrule
\multicolumn{1}{c|}{Backbone} &
  \multicolumn{1}{c|}{Method} &
  R@1 &
  R@2 &
  R@3 \\ \midrule
\multicolumn{1}{c|}{} &
  \multicolumn{1}{c|}{SEARLE} &
  14.4 &
  25.3 &
  35.4 \\
\multicolumn{1}{c|}{} &
  \multicolumn{1}{c|}{\cellcolor[HTML]{EFEFEF}CIReVL} &
  \cellcolor[HTML]{EFEFEF}15.8 &
  \cellcolor[HTML]{EFEFEF}26.8 &
  \cellcolor[HTML]{EFEFEF}36.8 \\
\multicolumn{1}{c|}{} &
  \multicolumn{1}{c|}{\cellcolor[HTML]{EFEFEF}OSrCIR} &
  \cellcolor[HTML]{EFEFEF}{\ul 17.4} &
  \cellcolor[HTML]{EFEFEF}{\ul 29.1} &
  \cellcolor[HTML]{EFEFEF}{\ul 39.0} \\
\multicolumn{1}{c|}{\multirow{-4}{*}{ViT-B/32}} &
  \multicolumn{1}{c|}{\cellcolor[HTML]{EFEFEF}\textbf{G-MIXER~(Ours)}} &
  \cellcolor[HTML]{EFEFEF}\textbf{18.3} &
  \cellcolor[HTML]{EFEFEF}\textbf{31.1} &
  \cellcolor[HTML]{EFEFEF}\textbf{41.7} \\ \midrule
\multicolumn{1}{c|}{} &
  \multicolumn{1}{c|}{SEARLE} &
  14.4 &
  25.3 &
  34.9 \\
\multicolumn{1}{c|}{} &
  \multicolumn{1}{c|}{PrediCIR} &
  16.6 &
  26.7 &
  35.8 \\
\multicolumn{1}{c|}{} &
  \multicolumn{1}{c|}{\cellcolor[HTML]{EFEFEF}CIReVL} &
  \cellcolor[HTML]{EFEFEF}15.9 &
  \cellcolor[HTML]{EFEFEF}27.1 &
  \cellcolor[HTML]{EFEFEF}33.8 \\
\multicolumn{1}{c|}{} &
  \multicolumn{1}{c|}{\cellcolor[HTML]{EFEFEF}OSrCIR} &
  \cellcolor[HTML]{EFEFEF}{\ul 17.9} &
  \cellcolor[HTML]{EFEFEF}{\ul 28.9} &
  \cellcolor[HTML]{EFEFEF}{\ul 38.7} \\
\multicolumn{1}{c|}{\multirow{-5}{*}{ViT-L/14}} &
  \multicolumn{1}{c|}{\cellcolor[HTML]{EFEFEF}\textbf{G-MIXER~(Ours)}} &
  \cellcolor[HTML]{EFEFEF}\textbf{19.9} &
  \cellcolor[HTML]{EFEFEF}\textbf{33.6} &
  \cellcolor[HTML]{EFEFEF}\textbf{43.9} \\ \midrule
\multicolumn{1}{c|}{} &
  \multicolumn{1}{c|}{PrediCIR} &
  13.7 &
  24.7 &
  33.6 \\
\multicolumn{1}{c|}{} &
  \multicolumn{1}{c|}{\cellcolor[HTML]{EFEFEF}CIReVL} &
  \cellcolor[HTML]{EFEFEF}17.4 &
  \cellcolor[HTML]{EFEFEF}29.8 &
  \cellcolor[HTML]{EFEFEF}39.5 \\
\multicolumn{1}{c|}{} &
  \multicolumn{1}{c|}{\cellcolor[HTML]{EFEFEF}OSrCIR} &
  \cellcolor[HTML]{EFEFEF}{\ul 19.6} &
  \cellcolor[HTML]{EFEFEF}{\ul 32.3} &
  \cellcolor[HTML]{EFEFEF}{\ul 42.5} \\
\multicolumn{1}{c|}{\multirow{-4}{*}{ViT-G/14}} &
  \multicolumn{1}{c|}{\cellcolor[HTML]{EFEFEF}\textbf{G-MIXER~(Ours)}} &
  \cellcolor[HTML]{EFEFEF}\textbf{20.2} &
  \cellcolor[HTML]{EFEFEF}\textbf{32.5} &
  \cellcolor[HTML]{EFEFEF}\textbf{42.6} \\ \bottomrule
\end{tabular}
\label{tab:gencis}
\caption{  \textbf{Comparison on GeneCIS Test Data.} Gray rows indicate training-free ZS-CIR methods. \textbf{Bold} and \underline{underlined} values denote the best and second-best results, respectively. }
\vspace{-2em}
\end{table}

\section{Experiments}
\subsection{Settings}
\textbf{Datasets.} 
We evaluate our method on four representative benchmarks for Zero-Shot Composed Image Retrieval (ZS-CIR): CIRCO~\cite{baldrati2023zero}, CIRR~\cite{liu2021image}, FashionIQ~\cite{wu2021fashion}, and GeneCIS~\cite{vaze2023genecis}. CIRCO and CIRR are composed of real-world images. CIRCO is the first CIR dataset that provides multiple ground-truths for each query, while CIRR includes an additional subset setting that assumes retrieval within a restricted image database. FashionIQ focuses on fine-grained clothing retrieval and consists of three subsets: Dress, Shirt, and Toptee.
GeneCIS is designed for conditional image similarity retrieval based on object–attribute reasoning, and it evaluates performance across four aspects: Focus Attribute, Change Attribute, Focus Object, and Change Object.

For evaluation metrics, we follow the common practice of the ZS-CIR literature.
We use mean Average Precision (mAP@K) for CIRCO, which has multiple ground-truths per query, and Recall@K (R@K) for CIRR, FashionIQ, and GeneCIS.

\noindent
\textbf{Baselines.} We compare our method with several ZS-CIR baselines, including both training-based and training-free approaches. The training-based methods rely on pseudo-token learning, while training-free methods utilize VLP and MLLMs.

\noindent
Training-based methods:
\begin{itemize}
    \item \textit{SEARLE}~\cite{baldrati2023zero}: Maps images into the text space by learning pseudo-word tokens through a dedicated network.
    \item \textit{PrediCIR}~\cite{tang2025missing}: Predicts the region of interest required for retrieval via a learned world model and converts cropped regions into pseudo tokens.
\end{itemize}

\noindent
Training-free methods
\begin{itemize}
    \item \textit{CIReVL}~\cite{karthik2023vision}: Uses a pre-trained captioner to describe the reference image and an LLM to combine it with the modification text into a target description.
    \item \textit{LDRE}~\cite{yang2024ldre}: Generates multiple target descriptions through diverse LLM reasoning and aggregates them via an ensemble strategy.
    \item \textit{OSrCIR}~\cite{tang2025reason}: Employs a MLLM with Reflective Chain-of-Thought reasoning to generate target descriptions.
\end{itemize}

\noindent
\textbf{Implementation Details.} 
For fair comparison, our implementation follows the overall experimental setup of OSrCIR~\cite{tang2025reason}. All experiments use the OpenAI GPT-4o model~\cite{achiam2023gpt} for caption and attribute generation, with the mixup ratio $\lambda$ varied within the range of 0.7 to 1.0 in increments of 0.05. The retrieval module, built in PyTorch~\cite{paszke2019pytorch} based on the codebase~\cite{tang2025reason} from, performs all computations on a single NVIDIA 4090 GPU.

\subsection{Quantitative Results}
Across all benchmarks, G-MIXER consistently outperforms both training-based and training-free baselines in Table 1, 2 and 3. In particular, on \textbf{CIRCO}~(ViT-L/14), G-MIXER achieves mAP@50 = 32.39\%, surpassing OSrCIR by +3.42\%. Even at a smaller retrieval scope (k=5), our method records 28.29\%, outperforming OSrCIR’s 23.87\% by +4.42\%. Also, on \textbf{CIRR}~(ViT-L/14), G-MIXER achieves mAP@50 = 77.69\%, outperforming OSrCIR’s 69.86\% by 7.83\%. This demonstrates that G-MIXER not only benefits from an expanded retrieval space but also maintains stable precision as the candidate pool grows, owing to the explicit attribute–based re-ranking process that effectively refines the retrieved results. 
On \textbf{FashionIQ}~(ViT-L/14), G-MIXER improves the average R@50 by +8.1\%, confirming its robustness in fine-grained retrieval scenarios involving subtle variations such as color, texture, and pattern. While PrediCIR shows competitive performance in certain fashion subsets due to its token-level specialization for garment attributes, G-MIXER clearly surpasses existing methods on broader benchmarks such as CIRR and CIRCO.
Notably, on CIRR (ViT-G/14), G-MIXER achieves 79.35\%, outperforming PrediCIR (77.9\%) and OSrCIR (77.33\%).
In addition, on the \textbf{GeneCIS}, G-MIXER consistently achieves the highest R@1–3 across all backbones, demonstrating its strong capability to reason about both object-level and attribute-level compositional changes.

\subsection{Qualitative Analysis}
The proposed G-MIXER expands the retrieval space through composed queries that incorporate implicit cues, effectively complementing the fine-grained details often missed by purely text-based reasoning. This approach is fundamentally different from conventional MLLM-based methods that explicitly convert all implicit information into textual descriptions for retrieval.

\begin{figure}[h!] 
\begin{center}
\includegraphics[width=1\linewidth]{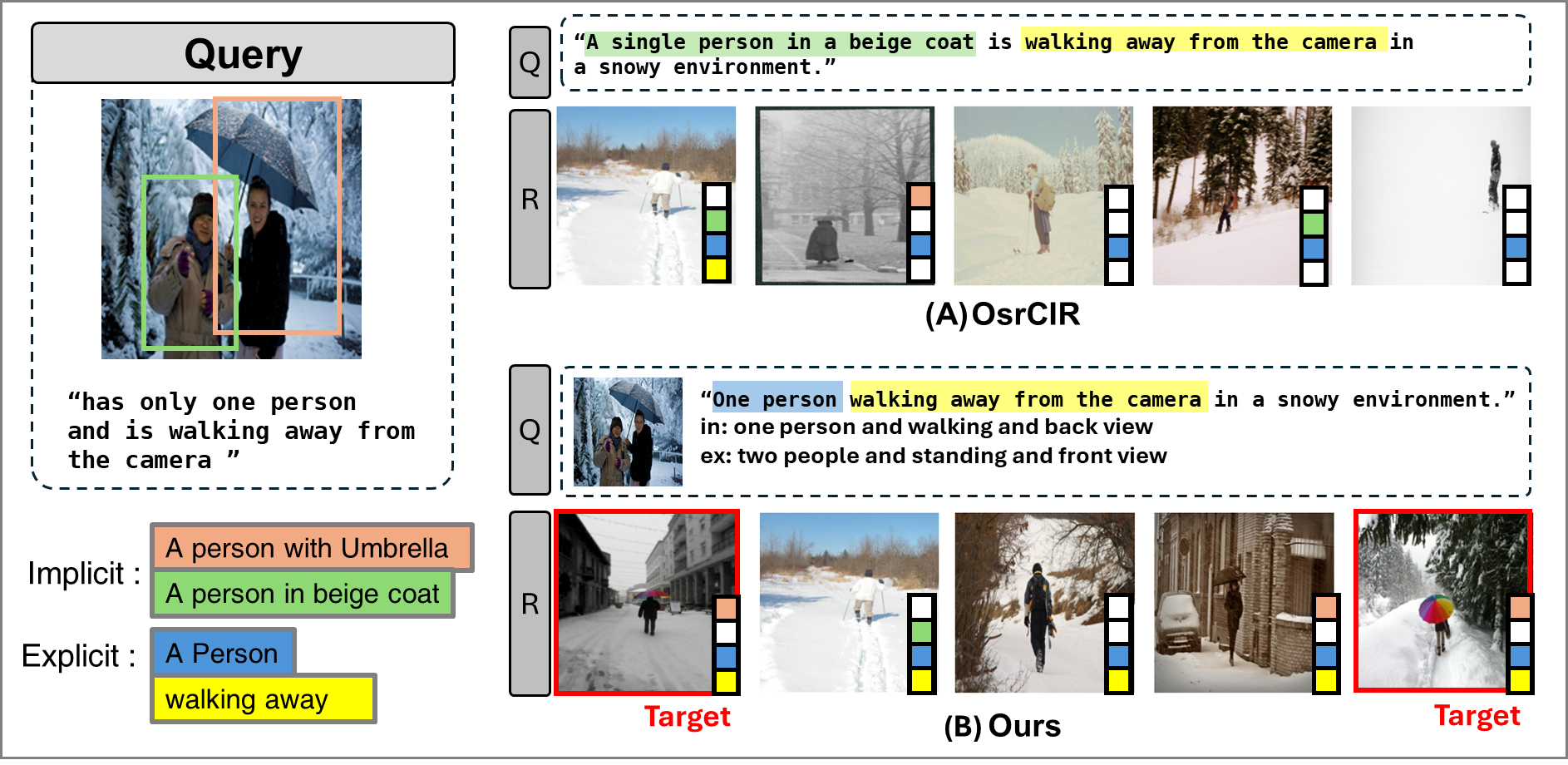}
\end{center}
\vspace{-1em}
\caption{  Qualitative results on CIRCO.  }
\label{fig:result1}
\end{figure}

As shown in Figure~\ref{fig:result1}, in the CIRCO dataset example, the modification text “has only one person and is walking away from the camera” omits the subject, making it unclear whether “one person” refers to the left or right individual. In the prior MLLM-based method (OSrCIR) (A), the LLM infers and generates the target caption “A single person in a beige coat walking away from the camera in a snowy environment.” However, the actual target image corresponds to the person holding an umbrella. Because the MLLM explicitly rewrites implicit details from the reference image into the target caption, the retrieval space becomes overly restricted, leading to reduced diversity in candidate results. In contrast, G-MIXER (B)
expands the retrieval range to include a broader set of compositional candidates, enabling effective retrieval of targets that require implicit contextual understanding.

\begin{figure}[h!] 
\begin{center}
\includegraphics[width=1\linewidth]{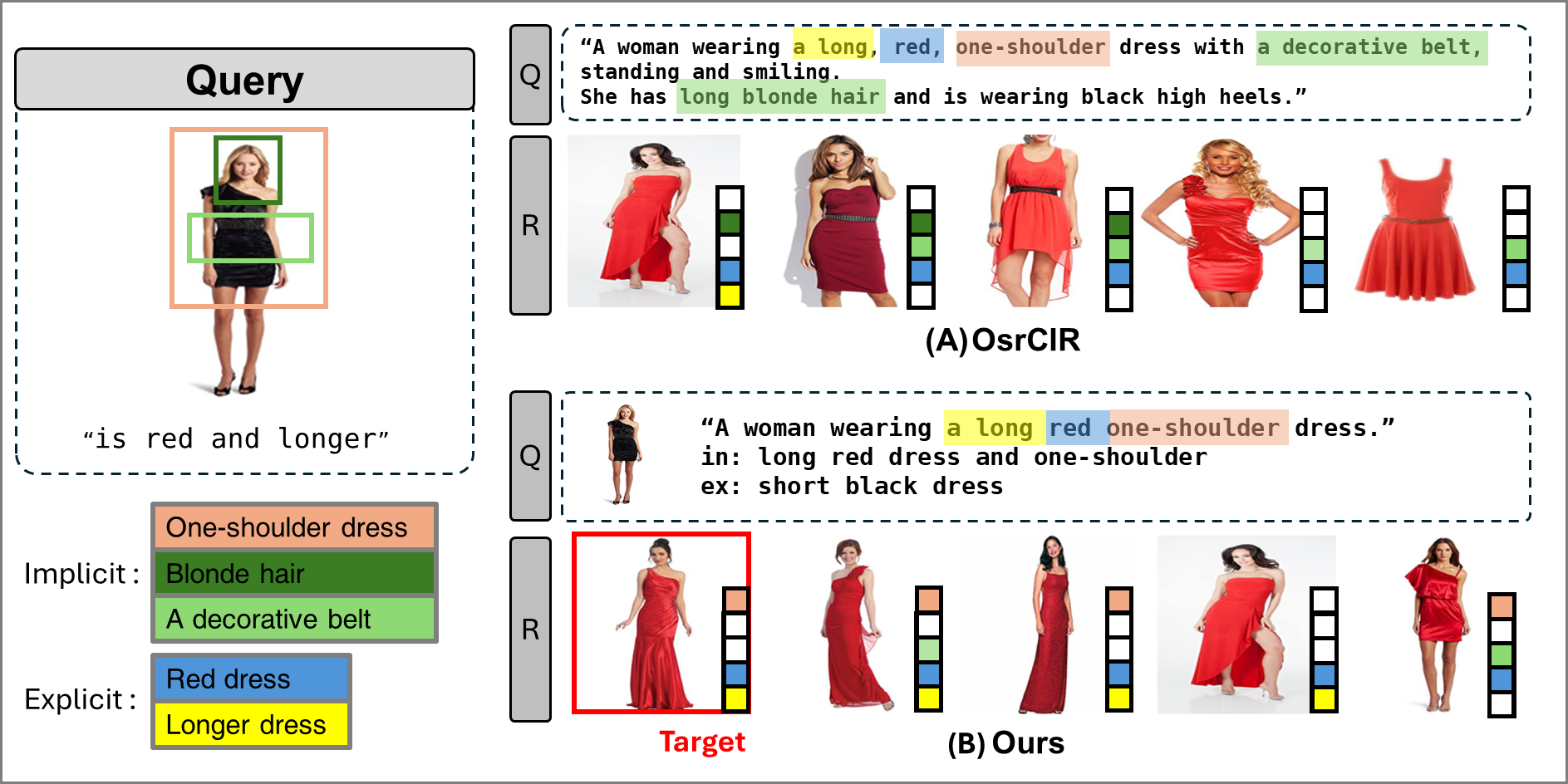}
\end{center}
\vspace{-1em}
\caption{  Qualitative results on Fashion-IQ dress.  }
\label{fig:result2}
\end{figure}

Figure~\ref{fig:result2} illustrates the comparison between the proposed G-MIXER and the prior MLLM-based method (OSrCIR) on the FashionIQ dataset. The existing method produces an overly detailed caption of the reference image, explicitly describing multiple implicit attributes (e.g., blonde hair, a decorative belt) that are not mentioned in the modification text. Such excessive explicitness introduces unnecessary constraints into the retrieval process, limiting the search space and lowering alignment with the true target. In contrast, G-MIXER performs re-ranking based solely on explicit modification cues (e.g., red dress, longer dress) while preserving implicit information implicitly through G-MIX. 
Consequently, G-MIXER accurately reflects essential attribute changes such as color, length, and shape, without being misled by non-essential factors like hair color or accessories. In summary, G-MIXER alleviates the limitations of over-explicit conversion in prior approaches by jointly achieving implicit information preservation and explicit information refinement. As a result, it delivers retrieval outcomes that balance both accuracy and diversity, particularly in fine-grained visual domains such as fashion.

\noindent
\subsection{Effectiveness and Efficiency Analysis. }

Our approach is a training-free ZS-CIR approach that requires no additional learning. The MLLM step takes about 0.6 seconds per query, accounting for approximately 97\% of the total time, while the remaining inference takes only 0.34 seconds, comparable to OSrCIR’s 0.32 seconds but achieving about 4\% points higher performance. Under the same experimental setting with pre-computed embeddings, the multiple mixup ratios introduce negligible computational overhead ($<$ 0.02s), as they only require additional cosine similarity computations that are efficiently parallelized on GPU.
Since both image and text embeddings can be pre-computed, the main cost lies in computing cosine similarities between the N queries and the candidate pool, keeping the overall GPU computation minimal while ensuring efficient and high-quality retrieval.


\subsection{Ablation Study and Analysis}
To verify the contribution of each component in the proposed G-MIXER, we conducted a series of ablation studies. All experiments were performed using the ViT-L/14 backbone, and the captions were generated with GPT-4o for a fair comparison. The results are summarized in Table~\ref{tab:ablation_component}


\begin{table}[h!]
\centering
\scriptsize
\renewcommand{\arraystretch}{1}
\setlength{\tabcolsep}{4.5pt}
\begin{tabular}{cccc|cc|cc|cc}
\toprule
\multicolumn{4}{c|}{\textbf{Components}} & \multicolumn{2}{c|}{\textbf{CIRCO}} & \multicolumn{2}{c|}{\textbf{CIRR}} & \multicolumn{2}{c}{\textbf{Fashion-IQ}} \\
 Mixup & $S_{\lambda}$  & $\Delta$ & $S_m$ & k=5 & k=10 & k=1 & k=5 & k=10 & k=50 \\ 
\midrule
\checkmark &  \checkmark &\checkmark & \checkmark 
& \textbf{28.29} & \textbf{29.04} &
\textbf{37.42} & \textbf{67.69} &
\textbf{41.92} & \textbf{62.47} \\ 
\midrule
\checkmark &  \checkmark   &  & \checkmark  & 22.49 & 23.77 & 34.87 & 67.06 & 40.61 & 61.05 \\ 
\checkmark &   &\checkmark & \checkmark& 16.43 & 17.47 & 27.40 & 53.23 & 32.36 & 57.38 \\ 
\checkmark &   &  & \checkmark&  11.80 & 12.97  &  28.34 &  55.08 & 30.76 & 56.00  \\

\checkmark &  \checkmark &  &  & 
11.32&13.24&25.59&55.71&36.82&56.71 \\
\checkmark & & \checkmark &   & 
12.28&13.34& 12.55 & 29.78  &24.94 & 53.02 \\
\checkmark & \checkmark  &\checkmark & & 
24.30 & 25.28 & 20.50 & 47.34 & 40.96 & 61.60 \\ 


\midrule
  &  \checkmark &\checkmark & \checkmark& 24.77 & 25.74 & 33.69 & 63.74 & 34.36 & 55.99 \\ 
\bottomrule
\end{tabular}
\caption{Ablation study on CIRCO, CIRR, and Fashion-IQ datasets using ViT-L/14.  }
\vspace{-2em}
\label{tab:ablation_component}
\end{table}

\subsubsection{ Effect of G-MIX }
\begin{figure}[h!] 
\begin{center}
\includegraphics[width=1\linewidth]{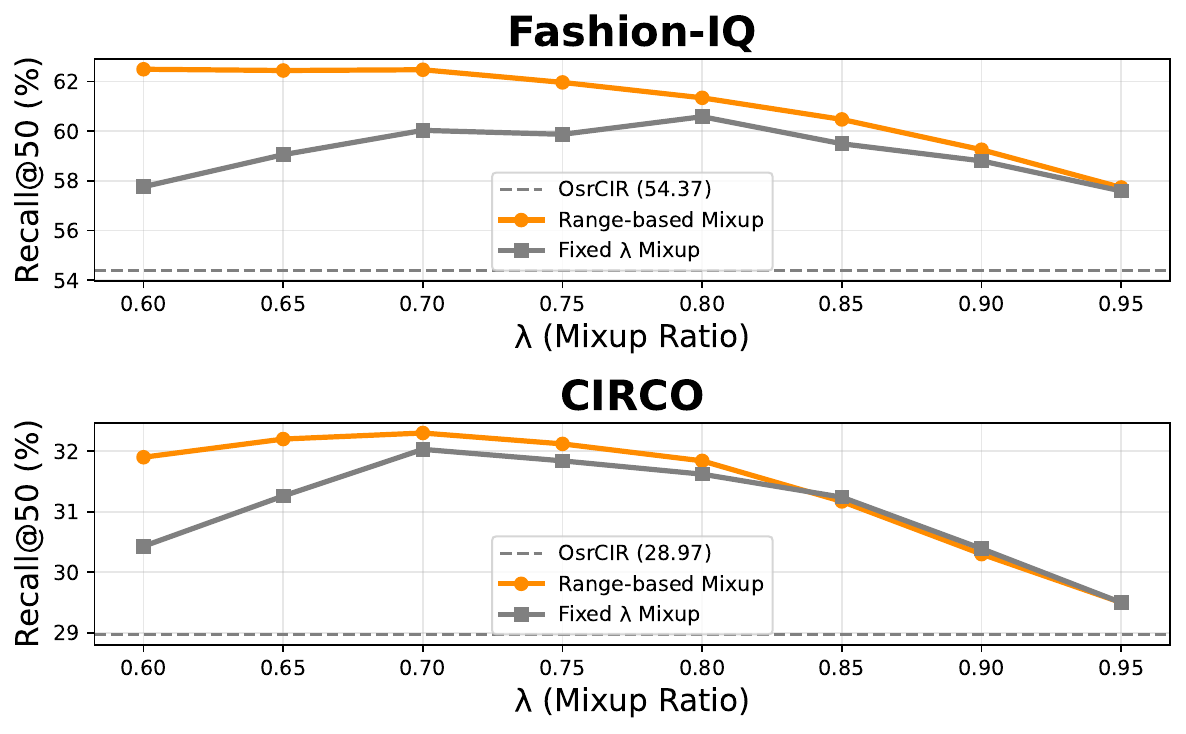}
\end{center}
\vspace{-2em}
\caption{Comparison between the proposed \textit{range-based} geodesic mixup (orange) and the \textit{fixed-ratio} mixup (gray). The range-based approach expands the retrieval space by applying a range of mixup ratios, while the fixed-ratio method uses a single static value.}
\vspace{-1em}
\label{fig:lamda}
\end{figure}

Unlike conventional mixup-based approaches that apply a single fixed mixup ratio to all queries, G-MIXER applies a range of mixup ratios ($\lambda$ range) to generate multiple composed features. These features are then integrated to form the first-stage candidate set, followed by a re-ranking process. 
Figure~\ref{fig:lamda} compares the range-based search with the conventional fixed-ratio setup, illustrating the effect of varying the starting value of $\lambda$. In the figure, the orange line represents the range-based search, while the gray line denotes the fixed-ratio search. For the range-based setup, the starting value of $\lambda$ was varied from 0.6 to 1.0 in increments of 0.05. (e.g., [0.6–1.0], [0.65–1.0], …, [0.95–1.0])

As shown, the fixed-ratio mixup occasionally achieves high performance at specific ratios but exhibits low consistency across datasets, as the optimal ratio differs for each domain. In contrast, the range-based mixup achieves the best performance when the starting $\lambda$ is between 0.6 and 0.7, while narrower ranges lead to performance degradation. This demonstrates that extending the mixup ratio to a range allows for stable and robust performance without the need for dataset-specific tuning. Therefore, employing a $\lambda$-range–based geodesic mixup is a key factor in expanding the retrieval scope and exploring the semantic space more effectively.

\begin{table}[t!]
\centering
\scriptsize
\renewcommand{\arraystretch}{1.15}
\setlength{\tabcolsep}{4.5pt}
\begin{tabular}{l|cc|cc|cc}
\toprule
\textbf{Methods} & 
\multicolumn{2}{c |}{\textbf{CIRCO}} &
\multicolumn{2}{c |}{\textbf{CIRR}} &
\multicolumn{2}{c}{\textbf{Fashion-IQ}} \\
& k=5 & k=10 &
k=1 & k=5 &
k=10 & k=50 \\ 
\midrule
\textbf{1. G-MIXER (GPT-4o)} 
& \textbf{28.29} & \textbf{29.04} &
\textbf{37.42} & \textbf{67.69} &
\textbf{41.92} & \textbf{62.47} \\ 
\midrule
\multicolumn{6}{l}{\textbf{ Different configuration of $\Delta$ in Eq.~\ref{eq:final} }} \\
2. $\Delta \rightarrow \Delta_{\text{in}} $ & 21.17 & 22.58 & 33.01 & 63.98 & 41.80 & 60.33 \\ 
3. $ \Delta \rightarrow \Delta_{\text{ex}} $ & 17.57 & 18.63 & 31.81 & 59.90 & 33.20 & 57.51 \\ 
\midrule
\multicolumn{6}{l}{\textbf{Impact of different MLLMs}} \\
5. GPT-turbo & 26.68 & 27.32 & 36.34 & 66.14 & 41.26 & 61.89 \\ 
6. GPT-4o-mini & 27.18  & 27.92 &  36.87 &  66.63 &  41.48 &  62.01 \\ 
\bottomrule
\end{tabular}
\caption{ Study of $\Delta$ variations in re-ranking score, and the impact of different MLLMs. }
\label{tab:ablation_reflectivecot}
\vspace{-1.5em}
\end{table}

\subsubsection{ Effect of ER }
\noindent
Explicit Semantic Re-ranking refines the first-stage candidate set by removing noisy candidate based on explicit information. 
The results in Table~\ref{tab:ablation_component} show the effect of excluding each component of the re-ranking score: $S_m$, $S_{\lambda}$, and $\Delta$ in Eq.~\ref{eq:delta}.
When both $S_{\lambda}$ and $\Delta$ were removed, performance dropped by \textbf{16.49\%}, showing that these two components work complementarily and are essential for achieving the full effectiveness of the proposed method.

Additionally, we conducted an ablation study to examine different configuration of the reward and penalty terms in the computation of $\Delta$ (Eq.~\ref{eq:delta}). 
\begin{equation}
\Delta \rightarrow \Delta_{\text{in}}
: {ReLU}(S_{\lambda} - S_{ex}) + {ReLU}(S_{in} - S_{\lambda})
\label{eq:delta_in}
\end{equation}
\vspace{-2em}
\begin{equation}
\Delta \rightarrow \Delta_{\text{ex}}
:- ReLU( S_{ex} - S_{\lambda} ) - ReLU(S_{\lambda}- S_{in}) 
\label{eq:delta_ex}
\end{equation}

For $\Delta_{\text{in}}$, penalizing cases where the similarity decreased was more effective than rewarding cases where it increased. This indicates that emphasizing similarity gains may cause overfitting to explicit cues, limiting the diversity of the expanded implicit representations. In contrast, for $\Delta_{\text{ex}}$, rewarding candidates that successfully removed excluded attributes performed better than applying penalties. This design allows the re-ranking process to selectively promote candidates that preserve the intended compositional modification while suppressing residual explicit noise from the geodesic mixup expansion.

\subsubsection{ Effects of the choice of MLLM }
We analyze the impact of the choice of MLLM among GPT-4o, GPT-4o-mini, and GPT-4o-turbo.
Even when using GPT-4o-mini, which is more efficient than GPT-4o, the performance drops by only about 1\%. The small variation across models implies that G-MIXER operates effectively regardless of the specific MLLM used.

 
\section{Conclusion}


We proposed G-MIXER, a training-free method for Zero-Shot Composed Image Retrieval (ZS-CIR) that expanded the retrieval scope through implicit semantic mixup and refined results using explicit cues.
By leveraging both implicit and explicit semantics to construct richer compositional representations of images and texts, G-MIXER effectively overcame the limitations of text-dominant approaches.
Experiments across multiple benchmarks demonstrated that G-MIXER consistently achieved superior retrieval performance over both training-based and training-free methods.




{
    \small
    \bibliographystyle{ieeenat_fullname}
    \bibliography{main}

@String(CVPR= {IEEE Conf. Comput. Vis. Pattern Recog.})

@String(AAAI = {AAAI})

@String(CVPR  = {CVPR})

@article{bordogna1993fuzzy,
  title={A fuzzy linguistic approach generalizing boolean information retrieval: A model and its evaluation},
  author={Bordogna, Gloria and Pasi, Gabriella},
  journal={Journal of the American Society for Information Science},
  volume={44},
  number={2},
  pages={70--82},
  year={1993},
  publisher={Wiley Online Library}
}

@article{chen2002region,
  title={A region-based fuzzy feature matching approach to content-based image retrieval},
  author={Chen, Yixin and Wang, James Ze},
  journal={IEEE transactions on pattern analysis and machine intelligence},
  volume={24},
  number={9},
  pages={1252--1267},
  year={2002},
  publisher={IEEE}
}

@article{tahani1976fuzzy,
  title={A fuzzy model of document retrieval systems},
  author={Tahani, Valiollah},
  journal={Information Processing \& Management},
  volume={12},
  number={3},
  pages={177--187},
  year={1976},
  publisher={Elsevier}
}

@inproceedings{tang2025reason,
  title={Reason-before-retrieve: One-stage reflective chain-of-thoughts for training-free zero-shot composed image retrieval},
  author={Tang, Yuanmin and Zhang, Jue and Qin, Xiaoting and Yu, Jing and Gou, Gaopeng and Xiong, Gang and Lin, Qingwei and Rajmohan, Saravan and Zhang, Dongmei and Wu, Qi},
  booktitle={Proceedings of the Computer Vision and Pattern Recognition Conference},
  pages={14400--14410},
  year={2025}
}

@inproceedings{vo2019composing,
  title={Composing text and image for image retrieval-an empirical odyssey},
  author={Vo, Nam and Jiang, Lu and Sun, Chen and Murphy, Kevin and Li, Li-Jia and Fei-Fei, Li and Hays, James},
  booktitle={Proceedings of the IEEE/CVF Conference on Computer Vision and Pattern Recognition (CVPR)},
  pages={6439--6448},
  year={2019}
}

@inproceedings{baldrati2022effective,
  title={Effective conditioned and composed image retrieval combining clip-based features},
  author={Baldrati, Alberto and Bertini, Marco and Uricchio, Tiberio and Del Bimbo, Alberto},
  booktitle={Proceedings of the IEEE/CVF conference on computer vision and pattern recognition},
  pages={21466--21474},
  year={2022}
}

@inproceedings{liu2021image,
  title={Image retrieval on real-life images with pre-trained vision-and-language models},
  author={Liu, Zheyuan and Rodriguez-Opazo, Cristian and Teney, Damien and Gould, Stephen},
  booktitle={Proceedings of the IEEE/CVF international conference on computer vision},
  pages={2125--2134},
  year={2021}
}

@inproceedings{baldrati2023zero,
  title={Zero-shot composed image retrieval with textual inversion},
  author={Baldrati, Alberto and Agnolucci, Lorenzo and Bertini, Marco and Del Bimbo, Alberto},
  booktitle={Proceedings of the IEEE/CVF International Conference on Computer Vision},
  pages={15338--15347},
  year={2023}
}

@inproceedings{saito2023pic2word,
  title={Pic2word: Mapping pictures to words for zero-shot composed image retrieval},
  author={Saito, Kuniaki and Sohn, Kihyuk and Zhang, Xiang and Li, Chun-Liang and Lee, Chen-Yu and Saenko, Kate and Pfister, Tomas},
  booktitle={Proceedings of the IEEE/CVF Conference on Computer Vision and Pattern Recognition},
  pages={19305--19314},
  year={2023}
}

@inproceedings{tang2024context,
  title={Context-i2w: Mapping images to context-dependent words for accurate zero-shot composed image retrieval},
  author={Tang, Yuanmin and Yu, Jing and Gai, Keke and Zhuang, Jiamin and Xiong, Gang and Hu, Yue and Wu, Qi},
  booktitle={Proceedings of the AAAI Conference on Artificial Intelligence},
  volume={38},
  number={6},
  pages={5180--5188},
  year={2024}
}

@inproceedings{radford2021learning,
  title={Learning transferable visual models from natural language supervision},
  author={Radford, Alec and Kim, Jong Wook and Hallacy, Chris and Ramesh, Aditya and Goh, Gabriel and Agarwal, Sandhini and Sastry, Girish and Askell, Amanda and Mishkin, Pamela and Clark, Jack and others},
  booktitle={International conference on machine learning},
  pages={8748--8763},
  year={2021},
  organization={PmLR}
}

@inproceedings{li2022blip,
  title={Blip: Bootstrapping language-image pre-training for unified vision-language understanding and generation},
  author={Li, Junnan and Li, Dongxu and Xiong, Caiming and Hoi, Steven},
  booktitle={International conference on machine learning},
  pages={12888--12900},
  year={2022},
  organization={PMLR}
}

@inproceedings{jang2024spherical,
  title={Spherical linear interpolation and text-anchoring for zero-shot composed image retrieval},
  author={Jang, Young Kyun and Huynh, Dat and Shah, Ashish and Chen, Wen-Kai and Lim, Ser-Nam},
  booktitle={European Conference on Computer Vision},
  pages={239--254},
  year={2024},
  organization={Springer}
}

@inproceedings{wu2021fashion,
  title={Fashion iq: A new dataset towards retrieving images by natural language feedback},
  author={Wu, Hui and Gao, Yupeng and Guo, Xiaoxiao and Al-Halah, Ziad and Rennie, Steven and Grauman, Kristen and Feris, Rogerio},
  booktitle={Proceedings of the IEEE/CVF Conference on computer vision and pattern recognition},
  pages={11307--11317},
  year={2021}
}

@inproceedings{vaze2023genecis,
  title={Genecis: A benchmark for general conditional image similarity},
  author={Vaze, Sagar and Carion, Nicolas and Misra, Ishan},
  booktitle={Proceedings of the IEEE/CVF Conference on Computer Vision and Pattern Recognition},
  pages={6862--6872},
  year={2023}
}

@article{tang2025missing,
  title={Missing Target-Relevant Information Prediction with World Model for Accurate Zero-Shot Composed Image Retrieval},
  author={Tang, Yuanmin and Yu, Jing and Gai, Keke and Zhuang, Jiamin and Xiong, Gang and Gou, Gaopeng and Wu, Qi},
  journal={arXiv preprint arXiv:2503.17109},
  year={2025}
}

@article{karthik2023vision,
  title={Vision-by-language for training-free compositional image retrieval},
  author={Karthik, Shyamgopal and Roth, Karsten and Mancini, Massimiliano and Akata, Zeynep},
  journal={arXiv preprint arXiv:2310.09291},
  year={2023}
}

@inproceedings{yang2024ldre,
  title={Ldre: Llm-based divergent reasoning and ensemble for zero-shot composed image retrieval},
  author={Yang, Zhenyu and Xue, Dizhan and Qian, Shengsheng and Dong, Weiming and Xu, Changsheng},
  booktitle={Proceedings of the 47th International ACM SIGIR conference on research and development in information retrieval},
  pages={80--90},
  year={2024}
}

@article{du2024image2sentence,
  title={Image2sentence based asymmetrical zero-shot composed image retrieval},
  author={Du, Yongchao and Wang, Min and Zhou, Wengang and Hui, Shuping and Li, Houqiang},
  journal={arXiv preprint arXiv:2403.01431},
  year={2024}
}

@article{gulanguage,
  title={Language-only Efficient Training of Zero-shot Composed Image Retrieval--Appendix--},
  author={Gu, Geonmo and Chun, Sanghyuk and Kim, Wonjae and Kang, Yoohoon and Yun, Sangdoo}
}

@inproceedings{psomas2025instance,
  title={Instance-Level Composed Image Retrieval},
  author={Psomas, Bill and Retsinas, George and Efthymiadis, Nikos and Filntisis, Panagiotis and Avrithis, Yannis and Maragos, Petros and Chum, Ondrej and Tolias, Giorgos},
  booktitle={The Thirty-ninth Annual Conference on Neural Information Processing Systems},
  year={2025}
}

@inproceedings{lin2024fine,
  title={Fine-grained textual inversion network for zero-shot composed image retrieval},
  author={Lin, Haoqiang and Wen, Haokun and Song, Xuemeng and Liu, Meng and Hu, Yupeng and Nie, Liqiang},
  booktitle={Proceedings of the 47th International ACM SIGIR Conference on Research and Development in Information Retrieval},
  pages={240--250},
  year={2024}
}

@inproceedings{suo2024knowledge,
  title={Knowledge-enhanced dual-stream zero-shot composed image retrieval},
  author={Suo, Yucheng and Ma, Fan and Zhu, Linchao and Yang, Yi},
  booktitle={Proceedings of the IEEE/CVF conference on computer vision and pattern recognition},
  pages={26951--26962},
  year={2024}
}

@article{agnolucci2025isearle,
  title={isearle: Improving textual inversion for zero-shot composed image retrieval},
  author={Agnolucci, Lorenzo and Baldrati, Alberto and Del Bimbo, Alberto and Bertini, Marco},
  journal={IEEE Transactions on Pattern Analysis and Machine Intelligence},
  year={2025},
  publisher={IEEE}
}

@article{alayrac2022flamingo,
  title={Flamingo: a visual language model for few-shot learning},
  author={Alayrac, Jean-Baptiste and Donahue, Jeff and Luc, Pauline and Miech, Antoine and Barr, Iain and Hasson, Yana and Lenc, Karel and Mensch, Arthur and Millican, Katherine and Reynolds, Malcolm and others},
  journal={Advances in neural information processing systems},
  volume={35},
  pages={23716--23736},
  year={2022}
}

@inproceedings{hummel2024egocvr,
  title={Egocvr: An egocentric benchmark for fine-grained composed video retrieval},
  author={Hummel, Thomas and Karthik, Shyamgopal and Georgescu, Mariana-Iuliana and Akata, Zeynep},
  booktitle={European Conference on Computer Vision},
  pages={1--17},
  year={2024},
  organization={Springer}
}

@article{jiang2023clip,
  title={From clip to dino: Visual encoders shout in multi-modal large language models},
  author={Jiang, Dongsheng and Liu, Yuchen and Liu, Songlin and Zhao, Jin'e and Zhang, Hao and Gao, Zhen and Zhang, Xiaopeng and Li, Jin and Xiong, Hongkai},
  journal={arXiv preprint arXiv:2310.08825},
  year={2023}
}

@article{liu2023visual,
  title={Visual instruction tuning},
  author={Liu, Haotian and Li, Chunyuan and Wu, Qingyang and Lee, Yong Jae},
  journal={Advances in neural information processing systems},
  volume={36},
  pages={34892--34916},
  year={2023}
}

@article{zhu2023minigpt,
  title={Minigpt-4: Enhancing vision-language understanding with advanced large language models},
  author={Zhu, Deyao and Chen, Jun and Shen, Xiaoqian and Li, Xiang and Elhoseiny, Mohamed},
  journal={arXiv preprint arXiv:2304.10592},
  year={2023}
}

@article{achiam2023gpt,
  title={Gpt-4 technical report},
  author={Achiam, Josh and Adler, Steven and Agarwal, Sandhini and Ahmad, Lama and Akkaya, Ilge and Aleman, Florencia Leoni and Almeida, Diogo and Altenschmidt, Janko and Altman, Sam and Anadkat, Shyamal and others},
  journal={arXiv preprint arXiv:2303.08774},
  year={2023}
}

@inproceedings{jia2021scaling,
  title={Scaling up visual and vision-language representation learning with noisy text supervision},
  author={Jia, Chao and Yang, Yinfei and Xia, Ye and Chen, Yi-Ting and Parekh, Zarana and Pham, Hieu and Le, Quoc and Sung, Yun-Hsuan and Li, Zhen and Duerig, Tom},
  booktitle={International conference on machine learning},
  pages={4904--4916},
  year={2021},
  organization={PMLR}
}

@article{yao2021filip,
  title={Filip: Fine-grained interactive language-image pre-training},
  author={Yao, Lewei and Huang, Runhui and Hou, Lu and Lu, Guansong and Niu, Minzhe and Xu, Hang and Liang, Xiaodan and Li, Zhenguo and Jiang, Xin and Xu, Chunjing},
  journal={arXiv preprint arXiv:2111.07783},
  year={2021}
}

@inproceedings{li2023blip,
  title={Blip-2: Bootstrapping language-image pre-training with frozen image encoders and large language models},
  author={Li, Junnan and Li, Dongxu and Savarese, Silvio and Hoi, Steven},
  booktitle={International conference on machine learning},
  pages={19730--19742},
  year={2023},
  organization={PMLR}
}

@article{bai2023qwen,
  title={Qwen-vl: A frontier large vision-language model with versatile abilities},
  author={Bai, Jinze and Bai, Shuai and Yang, Shusheng and Wang, Shijie and Tan, Sinan and Wang, Peng and Lin, Junyang and Zhou, Chang and Zhou, Jingren},
  journal={arXiv preprint arXiv:2308.12966},
  volume={1},
  number={2},
  pages={3},
  year={2023}
}

@inproceedings{liu2024improved,
  title={Improved baselines with visual instruction tuning},
  author={Liu, Haotian and Li, Chunyuan and Li, Yuheng and Lee, Yong Jae},
  booktitle={Proceedings of the IEEE/CVF conference on computer vision and pattern recognition},
  pages={26296--26306},
  year={2024}
}

@article{peng2023kosmos,
  title={Kosmos-2: Grounding multimodal large language models to the world},
  author={Peng, Zhiliang and Wang, Wenhui and Dong, Li and Hao, Yaru and Huang, Shaohan and Ma, Shuming and Wei, Furu},
  journal={arXiv preprint arXiv:2306.14824},
  year={2023}
}

@article{dai2023instructblip,
  title={Instructblip: Towards general-purpose vision-language models with instruction tuning},
  author={Dai, Wenliang and Li, Junnan and Li, Dongxu and Tiong, Anthony and Zhao, Junqi and Wang, Weisheng and Li, Boyang and Fung, Pascale N and Hoi, Steven},
  journal={Advances in neural information processing systems},
  volume={36},
  pages={49250--49267},
  year={2023}
}

@article{wu2024visionllm,
  title={Visionllm v2: An end-to-end generalist multimodal large language model for hundreds of vision-language tasks},
  author={Wu, Jiannan and Zhong, Muyan and Xing, Sen and Lai, Zeqiang and Liu, Zhaoyang and Chen, Zhe and Wang, Wenhai and Zhu, Xizhou and Lu, Lewei and Lu, Tong and others},
  journal={Advances in Neural Information Processing Systems},
  volume={37},
  pages={69925--69975},
  year={2024}
}

@article{paszke2019pytorch,
  title={Pytorch: An imperative style, high-performance deep learning library},
  author={Paszke, Adam and Gross, Sam and Massa, Francisco and Lerer, Adam and Bradbury, James and Chanan, Gregory and Killeen, Trevor and Lin, Zeming and Gimelshein, Natalia and Antiga, Luca and others},
  journal={Advances in neural information processing systems},
  volume={32},
  year={2019}
}
}


\end{document}